# Probabilistic Interpretations for MYCIN's Certainty Factors

David Heckerman

Medical Computer Science Group, TC135, Stanford University Medical Center

## Abstract

We propose a redefinition of the quantities called certainty factors used by MYCIN to manage uncertainty. This definition is based on the desiderata of certainty factors given in the original work. It is emphasized that this definition reflects the notion that certainty factors represent *changes* in belief about hypotheses given evidence as opposed to *absolute* beliefs. It is shown that this redefinition accommodates an infinite number of probabilistic interpretations. In constructing the interpretations, insight is provided into the assumptions made when propagating certainty factors through an inference net. For example, it is shown that all evidence which bears directly on a hypothesis must be conditionally independent on the hypothesis and its negation. Also, this analysis demonstrates that certainty factors must be elicited from experts in a precise fashion and a straightforward method for doing so is presented. Improvements in the methods currently used to propagate certainty factors through an inference net will also be suggested.

## Introduction

Many AI researchers have created *ad hoc* strategies for reasoning under uncertainty. They have invariably explained their approaches as necessary because of limitations of probability theory[1,2]. Many familiar with probability theory, however, argue that *ad hoc* methods for dealing with uncertainty are unnecessary and that probabilistic methods are sufficient in all cases[3,4,5,6,7]. In this paper, we will focus on the MYCIN certainty factor model[1], a purportedly *ad hoc* method for managing uncertainty that has seen wide use in rule-based expert systems. Shortliffe and Buchanan, creators of the model, justify the use of certainty factors by demonstrating system performance equal to that of experts in the field[8]. However, since they failed to provide a precise characterization of certainty factors, a good deal of mystery and confusion about the numbers has been generated. In this paper, we will clear up much of this mystery by showing that there are probabilistic interpretations for certainty factors. By demonstrating an equivalence between certainty factors and probabilities we will also lend support to the arguments that probability theory is sufficient for reasoning under uncertainty.

## Previous Work

This work is not the first to attempt to gain a better understanding of certainty factors. Adams[9] demonstrated several problems with the original definition of certainty factors. In an unpublished internal memo[10], Duda, Hart, Nilsson and Shortliffe proposed several new putative probabilistic interpretations but admittedly failed to acquire significant new insights into the certainty factor model. In fact, none of their interpretations fulfill the criteria for a probabilistic interpretation proposed below. Later, Duda demonstrated a strong similarity between the certainty factor model and the pseudo-probabilistic method used in the expert system PROSPECTOR to manage uncertainty[11]. We will consider this similarity in a new light. Finally, Hajek[12] examined the algebraic properties of certainty factors without attempting a probabilistic interpretation. The significance of his work within the current context will be discussed.

## MYCIN's Certainty Factors

Before examining probabilistic interpretations for MYCIN's certainty factors, we present a simplified overview of the quantities. MYCIN's knowledge is stored as *rules*. Rules are of the form IF *evidence* THEN *hypothesis*. In this paper, rules will be represented by

E ----------> H

where H is a hypothesis and E is evidence relating to the hypothesis. For example, a hypothesis might be that the organism *Staphylococcus* is infecting a patient while evidence for this hypothesis might be that the organism is gram-positive and grows in clumps.

In medicine, relationships between evidence and hypotheses are often uncertain. In order to accommodate these non-deterministic relationships, MYCIN uses certainty factors. To each rule, a certainty factor is attached which represents the *change* in belief about a hypothesis given some evidence. In this paper, we write

$$CF(H,E)$$
E ----------> H

Certainty factors range between -1 and 1. Positive numbers correspond to an *increase* in belief in a hypothesis while negative quantities correspond to a *decrease* in belief. Note that certainty factors do not correspond to measures of *absolute* belief. This has been a source of confusion with respect to certainty factors as well as other measures of uncertainty[13].

In MYCIN, it is possible that several pieces of evidence bear on the same hypothesis. It is also possible for a hypothesis to serve as evidence for another hypothesis. A interconnected set of hypotheses and evidence is called an *inference network*[14]. Figure 1 depicts one possible net. In the figure, $S_i$ is a statement which serves as either evidence or a hypothesis or sometimes both.

*This work was supported in part by the Josiah Macy, Jr. Foundation and the Henry J. Kaiser Family Foundation.

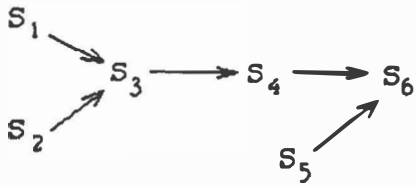

Figure 1: An inference net

Now let us examine the mechanisms used in MYCIN to propagate certainty factors through an inference net. In the certainty factor model, the propagation of belief is defined for two elementary inference nets. Propagation through a complex network is accomplished by decomposing the net into elementary structures where propagation is straightforward. The first of these elementary networks is diagramed below.

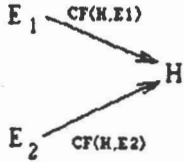

Here, two pieces of evidence bear on the same hypothesis. In MYCIN, these two rules are combined into an effective single rule with certainty factor $CF(H,E_1E_2)$ equal to

$$CF(H,E_1)+CF(H,E_2)-CF(H,E_1)CF(H,E_2) \text{ if } CF(H,E_1),CF(H,E_2)>0, \quad (1)$$

$$CF(H,E_1)+CF(H,E_2)+CF(H,E_1)CF(H,E_2) \text{ if } CF(H,E_1),CF(H,E_2)<0,$$

$$\frac{CF(H,E_1)+CF(H,E_2)}{1-\min(|CF(H,E_1)|,|CF(H,E_2)|)} \text{ otherwise.}$$

We will refer to this process as *parallel combination*.

In the second elementary network, a hypothesis serves as evidence for another hypothesis. Diagrammatically, we have

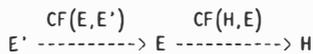

where E' is certain but E and H may not be certain. These two rules are combined into an effective single rule with a certainty factor denoted by $CF(H,E')$ where

$$CF(H,E') = CF(H,E) \max(0, CF(E,E')). \quad (2)$$

This will be called *sequential combination*.

To see how these combination formulas can be used to propagate certainty factors in more complicated situations, consider the following example:

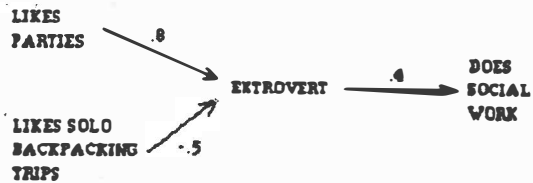

That is, if a person likes to go to parties, then the change in belief that he is an extrovert is .8. If he likes to make solo backpacking trips, the change in belief that he is an extrovert is -.5. Furthermore, if a person is an extrovert, then the change in belief that he participates in some type of social work is .4. To compute the change in belief that a person is an extrovert given that he likes to go to parties and also likes to make solo backpacking trips, we use the parallel combination function (1) with $CF(H,E_1) = .8$ and $CF(H,E_2) = -.5$ giving $CF(H,E_1E_2) = .8 - .5 / 1 - .5 = .6$. To compute the change in belief that a person does some type of social work given that he likes to go to parties and also likes to make solo backpacking trips, we use the sequential combination function (2) with $CF(E,E') = .6$ (the result of parallel combination) and $CF(H,E) = .4$ giving $CF(H,E') = (.6)(.4) = .24$.

This decompositional approach to propagation is straightforward when an inference net has a tree structure. However, MYCIN's inference net is not a tree and additional *ad hoc* procedures were introduced to handle complexities such as multiple paths from evidence to hypothesis. To simplify this discussion, we will only consider tree networks[*].

### Overview of interpretation development

As mentioned above, the certainty factor $CF(H,E)$ represents the *change* in belief of a hypothesis H based on the evidence E. In the original work[1], the creators of certainty factors presented this basic notion and then immediately proposed a *formal definition* of certainty factors in terms of probabilities. In somewhat simplified form and in our notation, Shortliffe and Buchanan gave as the *definition* of $CF(H,E)$:

$$CF(H,E) = \begin{cases} \dfrac{p(H|E) - p(H)}{1 - p(H)} & p(H|E) > p(H) \\[2ex] \dfrac{p(H|E) - p(H)}{p(H)} & p(H) > p(H|E) \end{cases} \quad (3)$$

Given this definition, however, they could not derive exact formulae for combining certainty factors (we will give one reason for this difficulty in a later section). Instead, they proposed "approximate" combination functions. They justified these functions by showing that they satisfied certain properties or "desiderata" which are consistent with the basic notion of certainty factors. For example, one of the desiderata is that parallel combination must not depend on the order in which evidence is presented. We will examine these properties in detail shortly.

Much of the efforts that have gone into trying to understand MYCIN's certainty factors have focused on problems with the original definition[9]. In this paper, we take a different approach; rather than define certainty factors in terms of probabilities, we define them in terms of the desiderata outlined in the original work. That is, we take the desiderata to be the defining *axioms* of certainty factors. We then look for combinations of probabilistic quantities in the spirit of (3) which are consistent with these axioms. These combinations are probabilistic interpretations for certainty factors defined in this new way. We use these interpretations to gain new insight into the use of certainty factors to manage uncertainty.

To proceed, it is important to make two distinctions. The first is the distinction between the combination functions used by MYCIN and the desiderata or defining axioms. To highlight this distinction, the symbol $\Delta$ or the phrase "certainty factor" will be used to refer to any quantity which satisfies the new definition of certainty factors while the term "CF" will refer to certainty factors that are combined with the MYCIN functions. The symbol $\Delta$ was chosen to emphasize that the quantity represents a belief *update*. A second distinction is made between certainty factors in general and particular probabilistic

---

[*]Pearl[15] gives a detailed discussion of belief propagation through more complicated networks

interpretations that may be ascribed to them. To highlight this distinction, a subscripted $\Delta$, e.g. $\Delta_1$, will refer to a particular probabilistic interpretation.

## Formal properties of certainty factors

We begin the construction of probabilistic interpretations for certainty factors by closely examining the desiderata or defining axioms. To understand these axioms, it is important to remember that $\Delta(H,E)$ represents the *change* in belief of hypothesis H given evidence E. In other words, $\Delta$ is a belief *update*. The defining axioms fall into three categories: (1) basic properties not related to the combination functions, (2) properties of parallel combination, and (3) properties of sequential combination. Since sequential combination is somewhat more complicated than parallel combination, we will first discuss categories one and two and return to category three later.

### Basic axioms

There are only two basic axioms of certainty factors. The first states that $\Delta$'s are real numbers in the interval [-1,1]. The second states that $\Delta$'s are ordered by the usual "<" relation on real numbers; that is, it makes sense to say that one certainty factor is less than another.

### Axioms of parallel combination

The first axiom of parallel combination was not made explicit in the original work. Consider two rules with the same hypothesis H

E --------> H

E' --------> H

It is possible that the update $\Delta(H,E)$ may depend on whether or not E' is known. In general, the update $\Delta(H,E)$ may depend on evidence for H already known. Therefore, certainty factors logically must have three arguments: H, E and e where e denotes prior evidence for H. However, the creators of certainty factors omitted the third argument, implicitly assuming that an update is independent of previous updates. This was done so that rules would retain a modular character[16]. Formally, the first axiom of parallel combination is

$$\Delta(H,E,e) = \Delta(H,E,\emptyset) \equiv \Delta(H,E) \quad (4)$$

for every e which does not entail E. This axiom will be called the modularity axiom. We will see later that it corresponds to the assumption of conditional independence.

As mentioned above, the parallel combination of two pieces of evidence $E_1$ and $E_2$ with respect to a hypothesis H is denoted $\Delta(H,E_1E_2)$. Given this notation, $\Delta$'s are required to be associative

$$\Delta(H,(E_1E_2)E_3) = \Delta(H,E_1(E_2E_3)) \quad (5)$$

and commutative

$$\Delta(H,E_1E_2) = \Delta(H,E_2E_1). \quad (6)$$

In other words, updating belief should not depend on how evidence is grouped or on the order in which it combined.

It is required that parallel combination preserves the ordering of updates. Suppose we have one update greater than another, that is $\Delta(H,E_1) > \Delta(H,E_2)$. When we combine each of these with a third update $\Delta(H,E_3)$, we want the two resulting updates to have the same relationship. Namely,

$$\text{if } \Delta(H,E_1) > \Delta(H,E_2) \text{ then } \Delta(H,E_1E_3) > \Delta(H,E_2E_3) \quad (7)$$

Another axiom of parallel combination states that 0 is a special certainty factor. In particular, if $\Delta(H,E)$ is 0, the evidence E does not change the belief in hypothesis H. This corresponds to a "no information" situation. In our notation,

$$\text{if } \Delta(H,E_1) = 0 \text{ then } \Delta(H,E_1E_2) = \Delta(H,E_2) \quad (8)$$

for all $E_2$. In other words, 0 is the identity element with respect to parallel combination.

The next property of certainty factors we will discuss was not considered a desideratum in the original work. Rather, it was shown to follow from the definition of the MYCIN parallel combination. We will upgrade this property to an axiom. Informally the axiom states that if $\Delta(H,E)$ corresponds to the *increase* in belief of H given E, then $-\Delta(H,E)$ should correspond to an equal and opposite *decrease* in belief. Formally,

$$\text{if } \Delta(H,E_1) = -\Delta(H,E_2) \text{ then } \Delta(H,E_1E_2) = 0 \quad (9)$$

That is, every certainty factor has an inverse with respect to parallel combination and it's inverse is just the certainty factor with opposite sign.

Finally, combination of "extreme" certainty factors must be consistent with the notion that a $\Delta(H,E)$ of 1 corresponds to the case where E proves H and a $\Delta(H,E)$ of -1 corresponds to the situation where H is disproved by E.

$$\text{if } \Delta(H,E_1) = 1 \text{ and } \Delta(H,E_2) \Leftrightarrow -1 \text{ then } \Delta(H,E_1E_2) = 1$$

$$\text{if } \Delta(H,E_1) = -1 \text{ and } \Delta(H,E_2) \Leftrightarrow 1 \text{ then } \Delta(H,E_1E_2) = -1 \quad (10)$$

$$\text{if } \Delta(H,E_1) = 1 \text{ and } \Delta(H,E_2) = -1 \text{ then } \Delta(H,E_1E_2)$$
$$\text{is undefined}.$$

The last case corresponds to conflicting evidence where $E_1$ proves H while $E_2$ disproves H.

## Requirements of a probabilistic interpretation

Intuitively, a probabilistic interpretation for certainty factors is a combination of probabilistic quantities, in the spirit of (3), which satisfies the defining axioms. In this section, we will formalize this notion.

First, we require that there is some function f such that

$$\Delta(H,E) = f(p(H|E), p(H)) \quad (11)$$

where $p(H|E)$ denotes the conditional probability of H given E, the *posterior* probability of H, and $p(H)$ denotes the probability of H before E is known, the *prior* probability. The only restrictions we place on f here is that it be a smooth function*. Later we will see that the axioms of certainty factors place much tighter restrictions on f. For later discussion, it is useful to state the above requirement with explicit reference to prior evidence e:

$$\Delta(H,E,e) = f(p(H|Ee), p(H|e)). \quad (12)$$

In addition to the above basic requirement, a probabilistic interpretation should be order preserving. In other words, suppose one update is greater than another. Given the same prior probability, the posterior probability corresponding to the first update should be greater than the posterior corresponding to the second. More formally,

$$\Delta(H,E_1) < \Delta(H,E_2) \text{ iff } p(H|E_1) < p(H|E_2) \quad (13)$$

with $p(H)$ fixed.

Finally, a probabilistic interpretation should be consistent with the notion that $\Delta(H,E) = 1$ when H follows from E and $\Delta(H,E) = -1$ when E disproves H.

$$\Delta(H,E) = 1 \quad \text{iff} \quad (p(H|E) = 1 \text{ and } p(H) \Leftrightarrow 0) \quad (14)$$

$$\Delta(H,E) = 0 \quad \text{iff} \quad (p(H|E) = 0 \text{ and } p(H) \Leftrightarrow 1)$$

---

*Precisely, the first derivative of f must exist and be continuous.

## The original interpretation of certainty factors

In this section, we examine the original definition (3) of certainty factors. Here, we view (3) as a probabilistic interpretation for certainty factors rather than a definition. To emphasize this we write

$$\Delta_{orig}(H,E) = \begin{cases} \dfrac{p(H|E) - p(H)}{1 - p(H)} & p(H|E) > p(H) \\ \\ \dfrac{p(H|E) - p(H)}{p(H)} & p(H) > p(H|E) \end{cases} \quad (15)$$

There are several problems with this putative interpretation. Here, it is shown that (15) dictates non-commutative parallel combination of evidence. In other words, this interpretation violates the commutativity axiom (6). This demonstration will help motivate the probabilistic interpretations presented later.

Consider two pieces of evidence which bear on the same hypothesis.

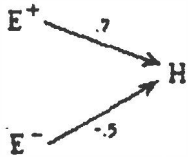

Suppose $p(H) = .4$. If we first update the probability of H with the positive evidence and then the negative evidence using (15), we get

$.7 = (p(H|E+) - .4) / (1 - .4) \implies p(H|E+) = .82$

$-.5 = (p(H|E+E-) - .82) / .82 \implies p(H|E+E-) = .41$

However, if we first update with the negative evidence and then the positive evidence, we get

$-.5 = (p(H|E-) - .4) / .4 \implies p(H|E-) = .2$

$.7 = (p(H|E-E+) - .2) / (1 - .2) \implies p(H|E-E+) = .76$

So we see that $p(H|E+E-) <> p(H|E-E+)$. In other words, the original interpretation leads to the unappealing result that the belief in a hypothesis is dependent on the order in which evidence for the hypothesis is considered.

## A probabilistic interpretation

The source of non-commutativity can be traced to the lack of symmetry in $\Delta_{orig}$. From (15) we see that positive evidence is measured relative to the distance between 1 and $p(H)$, while negative evidence is measured relative to the distance between 0 and $p(H)$ (see also figure 3 below). In order to avoid this asymmetry, it seems reasonable to map $p(H|E)$ and $p(H)$ into the interval $(-\infty, \infty)$. This gives some "elbow room" in which to combine $\Delta$'s without asymmetries. In fact, the obvious choice for a "certainty factor" in the infinite space is simply the distance between the mapped points of the posterior and prior probabilities. In this case, we can combine these certainty factors by adding them. However, there is now the problem that certainty factors in this space range from $-\infty$ to $\infty$. To fix this problem, we map these numbers into the interval $(-1, 1)$ with a function that preserves the $\Delta$ properties. This is shown diagrammatically in figure 2. Note that we are explicitly representing prior evidence e for reasons that will become clear

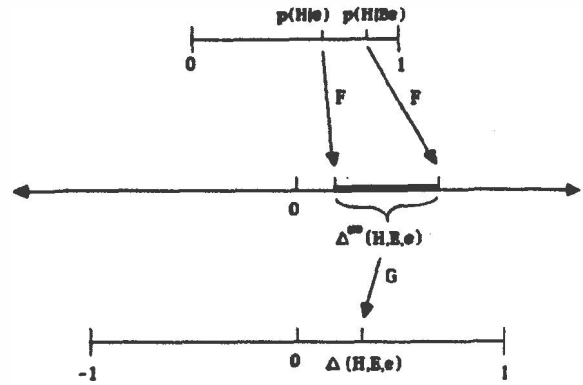

**Figure 2:** The mapping scheme for generating probabilistic interpretations

shortly. However, this is not necessary given the modularity axiom of $\Delta$'s, (4).

Referring to figure 2, the function F maps the posterior and prior probabilities in the interval (0,1) into $(-\infty, \infty)$. Certainty factors in the this infinite space, denoted $\Delta^{\infty}$, are then constructed by subtracting the mapped posterior from the mapped prior

$$\Delta^{\infty}(H,E,e) = F(p(H|Ee)) - F(p(H|e)) \quad (16)$$

Finally, the function G maps the $\Delta^{\infty}$'s into the interval $(-1, 1)$. That is,

$$\Delta(H,E,e) = G(\Delta^{\infty}(H,E,e)). \quad (17)$$

We cannot choose the functions F and G arbitrarily because the axioms of certainty factors and the requirements for a probabilistic interpretation must be respected. Later we will see that there is no freedom in the choice for F. In particular,

$$F(x) = \ln(x/1-x) \quad (18)$$

There is more freedom in the choice for G however. We will see that any G which satisfies

$G(x) = -G(-x)$

G monotonic increasing $\quad (19)$

$G(\infty) = 1$

will generate a probabilistic interpretation.

In this section, we will examine the probabilistic interpretation generated by

$$G_1(x) = (e^x - 1) / (e^x + 1). \quad (20)$$

It is easy to show that this choice for G satisfies (19). Combining (16) - (20), we get

$$\Delta_1(H,E,e) = \frac{p(H|Ee) - p(H|e)}{p(H|Ee) + p(H|e) - 2p(H|Ee)p(H|e)} \quad (21)$$

or, more simply,

$$\Delta_1(H,E) = \frac{p(H|E) - p(H)}{p(H|E) + p(H) - 2p(H|E)p(H)} \quad (22)$$

As mentioned earlier, we denote a particular probabilistic interpretation with a subscript. We also need to subscript the function G since any function satisfying (19) produces a probabilistic interpretation. There is no need to subscript $\Delta^{\infty}$ because F is fixed.

The parallel combination function for $\Delta_1$ can be derived from the combination function for $\Delta^\infty$. Recall that $\Delta^\infty$ was constructed such that it's parallel combination function is simply addition. That is,

$$\Delta^\infty(H, E_1E_2) = \Delta^\infty(H, E_1) + \Delta^\infty(H, E_2) \qquad (23)$$

This is also clear from (16). In combination with (17), this gives

$$G_1^{-1}(\Delta_1(H, E_1E_2)) = G_1^{-1}(\Delta_1(H, E_1)) + G_1^{-1}(\Delta_1(H, E_2)) \qquad (24)$$

and since

$$G_1^{-1} = \ln(1+x/1-x) \qquad (25)$$

we get

$$\frac{1 + \Delta_1(H,E_1E_2)}{1 - \Delta_1(H,E_1E_2)} = \frac{1 + \Delta_1(H,E_1)}{1 - \Delta_1(H,E_1)} \cdot \frac{1 + \Delta_1(H,E_2)}{1 - \Delta_1(H,E_2)} \qquad (26)$$

It is straightforward to show that $\Delta_1$ with the combination function (26) satisfies the basic axioms and the axioms of parallel combination as well as the requirements of a probabilistic interpretation. We say that $\Delta_1$ *is a probabilistic interpretation for certainty factors*.

The numerator of $\Delta_1$ clearly demonstrates that it is a probability update. In fact, $\Delta_1$ is not too different from the original definition of certainty factors. The numerators are the same while $\Delta_1$ seems to have a more complicated denominator. To get a better feel for $\Delta_1$, we can rewrite (22) as

$$\Delta_1(H,E) = \frac{p(H|E) - p(H)}{p(H)(1 - p(H|E)) + p(H|E)(1 - p(H))} \qquad (27)$$

Figure 3 depicts $\Delta_{orig}$ and $\Delta_1$ geometrically. Note the symmetry in the denominator of $\Delta_1$ that is lacking in the original interpretation.

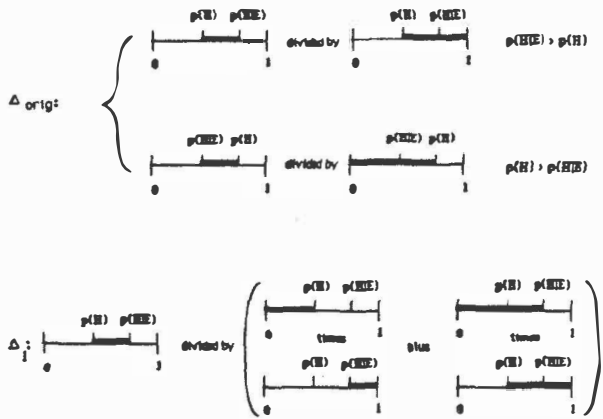

**Figure 3:** A geometrical comparison of $\Delta_{orig}$ and $\Delta_1$

### A comparison of the combination functions

It is interesting to compare the parallel combination function we have derived for $\Delta_1$, (26), with the combination function used in MYCIN, (1). Although both functions obey the axioms of parallel combination, there is no reason to expect that the two functions are equivalent. It turns out, however, that they are close. In figure 4, the two functions are plotted for $CF(H,E_1) = \Delta_1(H,E_1) = 0.5$. The two curves never differ by more than a few percent. It can easily be shown that the two combination functions differ the most when

$$CF(H,E_1) = \Delta_1(H,E_1) \text{ approaches } 1$$

and

$$CF(H,E_2) = \Delta_1(H,E_2) = -1 + (1 - CF(H,E_1))/2$$

In this case, $CF(H,E_1E_2)$ approaches $-1/2$ while $\Delta(H,E_1E_2)$ approaches $-1/3$.

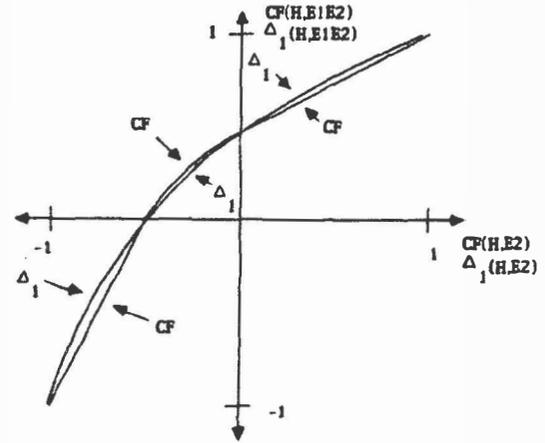

**Figure 4:** A comparison of the parallel combination function used in MYCIN and the parallel combination function for $\Delta_1$

### Relationship to Bayes' theorem

In this section, we will show that the probabilistic interpretation $\Delta_1$ is consistent with Bayes' theorem under the assumption of conditional independence. To begin, consider Bayes' theorem for updating the probability of a hypothesis H given prior evidence $e$ and new evidence E:

$$p(H|Ee) = \frac{p(E|He)p(H|e)}{p(E|e)} \qquad (28)$$

The corresponding formula for the negation of the hypothesis, $\sim H$, is

$$p(\sim H|Ee) = \frac{p(E|\sim He)p(\sim H|e)}{p(E|e)} \qquad (29)$$

Dividing (28) by (29) we get

$$\frac{p(H|Ee)}{p(\sim H|Ee)} = \frac{p(E|He)}{p(E|\sim He)} \cdot \frac{p(H|e)}{p(\sim H|e)} \qquad (30)$$

Now the odds of some event x, denoted $O(x)$, is just

$$O(x) = p(x)/(1 - p(x))$$

so that (30) can be written as

$$O(H|Ee) = \frac{p(E|He)}{p(E|\sim He)} O(H|e) \qquad (31)$$

Note that this form of Bayes' theorem was derived without assuming conditional independence of evidence.

Given (16) and (18) we can write

$$\Delta^\infty(H,E,e) = \ln(O(H|Ee)/O(H|e)) \qquad (32)$$

or using (31),

$$\Delta^\infty(H,E,e) = \ln(p(E|He)/p(E|\sim He)) \qquad (33)$$

Furthermore, with $G_1$ given by (20) we see that

$$\Delta_1(H,E,e) = (p(E|He)/p(E|\sim He) - 1)/(p(E|He)/p(E|\sim He) + 1) \quad (34)$$

or

$$p(E|He)/p(E|\sim He) = (\Delta_1(H,E,e) + 1)/(\Delta_1(H,E,e) - 1). \quad (35)$$

Now assume that E and e are conditionally independent given H and $\sim$H. That is

$$p(E|He) = p(E|H) \text{ and } p(E|\sim He) = p(E|\sim H) \quad (36)$$

for all e which does not entail E. Given this assumption, we see that the modularity axiom for $\Delta$'s, (4), is satisfied. Also, under assumption (36) we can write Bayes' theorem for two pieces of evidence as

$$\frac{p(H|E_1E_2)}{p(\sim H|E_1E_2)} = \frac{p(E_1E_2|H)}{p(E_1E_2|\sim H)} \frac{p(H)}{p(\sim H)} = \frac{p(E_1|H)}{p(E_1|\sim H)} \frac{p(E_2|H)}{p(E_2|\sim H)} \frac{p(H)}{p(\sim H)}. \quad (37)$$

Substituting (35) into (37) then gives the parallel combination function for $\Delta_1$, (26).

Therefore, the probabilistic interpretation $\Delta_1$ for certainty factors is consistent with Bayes' theorem with the assumption of conditional independence of evidence on H and its negation. Later we will see that the axioms of certainty factors and the requirements of a probabilistic interpretation for certainty factors mandates the conditional independence assumption.

### Relationship to the likelihood ratio

The quantity $p(E|H)/p(E|\sim H)$ is commonly called the *likelihood ratio* and written $\lambda(H,E)$. With this shorthand, we can write (31) as

$$\frac{O(H|E)}{O(H)} = \lambda(H,E). \quad (38)$$

This equation is often called the odds-likelihood form of Bayes' theorem. Under the assumption of conditional independence, the parallel combination function for $\lambda(H,E)$ is simple multiplication:

$$\lambda(H,E_1E_2) = \lambda(H,E_1) \lambda(H,E_2). \quad (39)$$

This follows from the definition of lambda, (38) and (37). The parallel combination function for the logarithm of the likelihood ratio, $\ln[\lambda(H,E)]$, which is identical to $\Delta^\infty(H,E)$ is just addition. That is,

$$\ln[\lambda(H,E_1E_2)] = \ln[\lambda(H,E_1)] + \ln[\lambda(H,E_2)]. \quad (40)$$

This desirable property of the log-likelihood ratio was first noted by Peirce in 1878[17]. He called the quantity $\ln[\lambda(H,E)]$ the "weight of evidence" for a hypothesis and argued that it is a natural form for the expression and processing of beliefs. This property of the log-likelihood ratio was also discovered independently by Turing[18], Good[19, 20] who also called them "weights of evidence" and Minsky[21]. The additive property of the log-likelihood ratio also reveals a striking resemblance to the *ad hoc* scoring scheme of INTERNIST-1. Indeed, it is not surprizing that such a simple and powerful method for reasoning with uncertainty has been "discovered" many times.

The likelihood ratio is used explicitly in several experts systems. For example, the GLASGOW DYSPEPSIA system uses $\ln\lambda(H,E)$[22] and PROSPECTOR uses $\lambda(H,E)$[14]. PROSPECTOR is of special interest because Duda, one of its designers, noticed a similarity between $\lambda(H,E)$ and certainty factors[11, 23]. In particular, he showed that the combination function for the quantity $\lambda(H,E)-1/\lambda(H,E)+1$, which he called $C(H,E)$, is quite similar to the MYCIN parallel combination function (1). From equation (34), we see that $C(H,E)$ is equal to $\Delta_1(H,E)$ and so the combination function for C is just (26). Duda considered the combination function for C to be an approximation of the combination function for CF. The concepts developed in this paper suggest that the one is no "better" than the other; both satisfy the axioms of parallel combination.

### Other probabilistic interpretations

The mapping scheme developed above can be used to create an infinite number of valid interpretations. Recall that any G which satisfies (19) will generate a probabilistic interpretation. One legitimate choice for G is of special interest. It is easy to show that the map

$$G(x) = \begin{cases} \ln(1/1-x) & x >= 0 \\ \ln(1+x) & x <= 0 \end{cases}$$

corresponding to the probabilistic interpretation

$$\Delta_{CF}(H,E) = \begin{cases} \dfrac{p(H|E) - p(H)}{p(H|E)(1 - p(H))} & p(H|E) > p(H) \\[2ex] \dfrac{p(H|E) - p(H)}{p(H)(1 - p(H|E))} & p(H) > p(H|E) \end{cases}$$

gives the MYCIN parallel combination function *exactly*.* That is, there is a probabilistic interpretation for certainty factors which yields the combination function used by MYCIN. We will not dwell on this mapping since we have shown that differences between the $\Delta_1$ and MYCIN parallel combination functions are small. Also, the latter mapping is more difficult to analyze and work with.

Consider the following set of choices for G.

$$G_n(x) = \frac{e^{x^{1/n}} - 1}{e^{x^{1/n}} + 1} \quad n = 1, 2, 3, \ldots$$

It is easy to check that each $G_n$ produces a valid probabilistic interpretation

$$\Delta_n(H,E) = \frac{(O(H|E)/O(H))^{1/n} - 1}{(O(H|E)/O(H))^{1/n} + 1}. \quad (41)$$

The parallel combination function corresponding to $G_n$ is given by*

$$\{\ln[1+\Delta_n(H,E_1E_2)/1-\Delta_n(H,E_1E_2)]\}^n = \{\ln[1+\Delta_n(H,E_1)/1-\Delta_n(H,E_1)]\}^n \quad (42)$$
$$+ \{\ln[1+\Delta_n(H,E_2)/1-\Delta_n(H,E_2)]\}^n.$$

As n approaches infinity, the combination function tends to select the certainty factor with the largest absolute value. Thus, the combination function can take on different qualitative behaviors depending on the probabilistic interpretation.* In the next section, we will examine an important implication of this fact. Here, we mention that (42) provides an explanation for a concern that the creators of certainty factors had about the behavior of their parallel combination function (1). They felt that combinations of positive evidence approached certainty (CF=1) too quickly[1]. Perhaps they had the probabilistic interpretation $\Delta_n$ (n large) in mind.

---

* This was first noted in a paper by Hajek[12].

* This example is given in a paper by Spiegelhalter[6].

* We will see later that G affects the sequential combination function as well.

### Eliciting certainty factors from experts

Certainty factors are often elicited from experts with rather vague phrases like: "Given E how strongly do you believe in H?" or "How strongly does E confirm H?." For example, these phrases were used during the construction of early versions of the MYCIN knowledge base[24]. Undoubtedly, the lack of a clear operational meaning for certainty factors is responsible for the use of imprecise phrases such as these. Moreover, such imprecision is troubling in light of the fact that different probabilistic interpretations can dictate parallel combination functions with qualitatively different behaviors. However, once certainty factors are given a probabilistic interpretation, it is a straightforward task to elicit them precisely.

Equations (32) and (33) show that the ratio $O(H|E)/O(H)$ or the equivalent ratio $p(E|H)/p(E|\sim H)$ is sufficient to determine $\Delta$ for any probabilistic interpretation. This suggests two methods for eliciting certainty factors. The first is to elicit $p(E|H)$ and $p(E|\sim H)$ separately and then use (34) to calculate $\Delta(H,E)$. The second is to elicit $p(H|E)$ assuming some $p(H)$ and then use (22). The second method has an interesting special case. If we let $p(H) = 1/2$, (22) gives

$$\Delta(H,E) = 2p(H|E) - 1 \qquad (43)$$

That is, suppose we ask the expert to imagine that the prior probability of some hypothesis is 1/2. Then we ask him for the posterior probability of the hypothesis given a piece of evidence. (43) states that this task of "starting" at 1/2 and giving a posterior probability is equivalent to the task of "starting" at 0 and giving a certainty factor in [-1,1]. This is shown diagrammatically in figure 5.

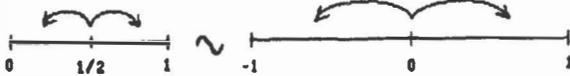

**Figure 5:** A correspondence between eliciting probabilities and eliciting certainty factors

It should be mentioned that eliciting probabilities is not a trivial task. Errors of bias are a major problem. For example, people tend to anchor on prior probabilities[25] and tend to be conservative near 0 and 1[26] (i.e., when probabilities are near one, they are underestimated and when probabilities are near zero, they are overestimated). There are techniques for overcoming biases[25, 27, 28, 28] but such a discussion is not appropriate here.

### Completeness of the mapping scheme

Earlier, we saw that it is possible to generate an infinite number of probabilistic interpretations for certainty factors with the map F given by (18) and the map G defined by (17). In this section, we show that this mapping scheme can generate *all* probabilistic interpretations. In other words, given any quantity which satisfies the basic axioms of certainty factors and the axioms of parallel combination as well as the requirements for a probabilistic interpretation, there is an F and G which generates it.

Consider the second half of the mapping scheme, (17). It turns out that certainty factors, as defined by the axioms, form a mathematical structure called an ordered commutative group.

This was first noted by Hajek[12]. A consequence of this fact is that given any quantity $\Delta$ and associated parallel combination function which satisfies the axioms, there is a function $\gamma$ such that

$$\gamma(\Delta(H,E_1 E_2)) = \gamma(\Delta(H,E_1)) + \gamma(\Delta(H,E_2))$$

where $\gamma$ must satisfy

$$\gamma(x) = -\gamma(-x)$$

$\gamma$ monotonic increasing $\qquad (44)$

$$\gamma(1) = \infty$$

In other words, $\Delta(H,E_1)$ and $\Delta(H,E_2)$ can combined by first mapping them to another space with the function $\gamma$, adding them in this new space, and then mapping the sum back to the original space with $\gamma^{-1}$. However, $\gamma^{-1}$ corresponds to what we have been calling G and $\gamma(\Delta(H,E))$ is just $\Delta^\infty(H,E)$. Therefore, given a quantity which satisfies the axioms of certainty factors, we will be able to find a G such that (17) and (23) are satisfied. The requirements on choices for G, (19), follow directly from (44).

Now consider the first half of the mapping scheme, (16). First note that the function G has an inverse because it is monotonic. Therefore, we can rewrite the first requirement for a probabilistic interpretation, (12), in terms of $\Delta^\infty$. That is, we require that there is some function f such that

$$\Delta^\infty(H,E,e) = f(p(H|Ee), p(H|e)). \qquad (45)$$

But (45) and (23) together imply that

$$f(p(H|E_1 E_2),p(H)) = f(p(H|E_1 E_2),p(H|E_1))+f(p(H|E_1),p(H)).$$

In other words, we know f satisfies

$$f(x,z) = f(x,y) + f(y,z)$$

for all x, y, and z between 0 and 1. Since f is smooth, there is some function $\phi$ such that

$$f(x,y) = \phi(y) - \phi(x).$$

However, it is clear from (16) and (45) that the function $\phi$ is just what we have been calling F. In other words, the relationship between $p(H|E)$ and $p(H)$ and $\Delta^\infty$ must be given by (16). Combining this result with the above we see that the mapping scheme, (16) and (17), can generate all probabilistic interpretations.

### The assumption of conditional independence

In the appendix, constraints for the function F similar to those for G in the previous section are derived. These constraints are intimately related to the assumption of conditional independence, (36). In fact, is shown that if the desiderata of parallel combination are to be respected, the only reasonable assumption to make about the dependencies among pieces of evidence which bear on a hypothesis is that of conditional independence.

In this section, we examine two important restrictions placed on the use of certainty factors by the assumption of conditional independence.[*] The first restriction is that certainty factors cannot be used to manage uncertainty in an inference net when any set of mutually exclusive and exhaustive hypotheses in the net has more than two elements. To see this, suppose $H_1$, $H_2$, and $H_3$ are mutually exclusive and exhaustive. A problem arises when we assume that evidence is conditionally independent on the $H_i$'s and their negation. For example,

---

[*] In a forthcoming paper, we will show that certainty factors can be generalized in such a way that the highly restrictive assumptions discussed in this section can be avoided.

suppose evidence is conditionally independent on $\sim H_3$. That is,

$$p(E|\sim H_3 e_1) = p(E|\sim H_3 e_2)$$

Now suppose that $e_1$ disproves $H_2$ and $e_2$ disproves $H_1$. Assuming that evidence is independent given $H_1$ and $H_2$, we get

$$p(E|H_1) = p(E|H_1 e_1) = p(E|H_2 e_2) = p(E|H_2)$$

which is not true in general.

The second restriction concerns two other types of combination defined in the certainty factor model. Those familiar with MYCIN may recall that the system uses "fuzzy set operators" to handle conjunctions and disjunctions of evidence in a rule premise. It is not very difficult to show that these functions violate the assumption of conditional independence. Therefore, these combinations must be avoided in order to adhere to the desiderata.

### Sequential combination

Now let us turn our attention to sequential combination. Recall, that one hypothesis can serve as evidence for another. Therefore, evidence for a hypothesis may itself be uncertain. As mentioned earlier, the simplest case can be represented as follows

$$\begin{array}{cc} \Delta(E,E') & \Delta(H,E) \\ E' \text{---------} > E \text{---------} > H \end{array}$$

where $E'$ is certain and $E$ and $H$ may not be certain. We are interested in constructing the single rule

$$E' \text{---------} > H$$

with certainty factor $\Delta(H,E')$.

What properties are required of sequential combination? Unfortunately, no desiderata concerning sequential combination were given in the original work so a set of axioms that seem simple and intuitively appealing are introduced.

The first three axioms concern the value of $\Delta(H,E')$ when when $\Delta(E,E')$ takes on the special values -1, 0, and 1. We require that if $\Delta(E,E') = 1$ then $\Delta(H,E') = \Delta(H,E)$,

if $\Delta(E,E') = -1$ then $\Delta(H,E') = \Delta(H,\sim E)$ and  (46)

if $\Delta(E,E') = 0$ then $\Delta(H,E') = 0$.

In the first case, $E'$ proves $E$ with certainty, in the second, $E'$ disproves $E$ with certainty and in the third, $E'$ says nothing about $E$. In the second case, it is clear that another certainty factor, $\Delta(H,\sim E)$, must be associated with every IF-THEN rule. That is, it is necessary to know the update for $H$ when $E$ is false. From (2) we see that in MYCIN, CF(H,E') is 0 whenever CF(E,E') = -1. Thus, MYCIN is implicitly assuming CF(H,$\sim$E) = 0 for every rule. We will return to this point shortly.

The fourth axiom states that $\Delta(H,E')$ is a function of only the certainty factors $\Delta(H,E)$, $\Delta(H,\sim E)$ and $\Delta(E,E')$. Informally, we want the net update to depend only on some combination of the individual rule updates. The final requirement of the sequential combination function is that $\Delta(H,E')$ is monotonic with respect to $\Delta(E,E')$.

Before doing a formal derivation, let us consider sequential updating on more intuitive grounds. Consider the following expression which follows from the definition of conditional probability.

$$p(H|E') = p(H|E,E')p(E|E') + p(H|\sim E,E')p(\sim E|E') \quad (47)$$

In order to satisfy the fourth axiom of sequential propagation, we assume

$$p(H|E,E') = p(H|E) \quad \text{and} \quad p(H|\sim E,E') = p(H|\sim E).$$

In this case, (47) becomes

$$p(H|E') = p(H|E)p(E|E') + p(H|\sim E)p(\sim E|E') \quad (48)$$

This formula is represented graphically in figure 6.

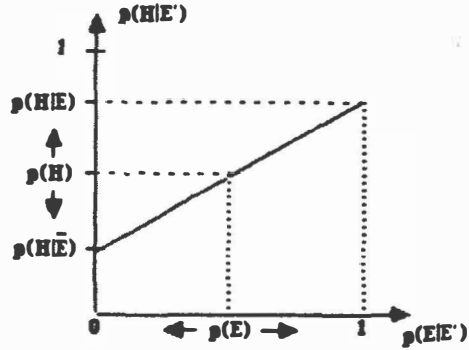

**Figure 6:** Sequential updating in terms of absolute probabilities

Before we know $E'$, $H$ and $E$ are at their prior values, $p(H)$ and $p(E)$, respectively.* The evidence $E'$ then updates the probability of $E$ to $p(E|E')$ which changes the probability of $H$ to $p(H|E')$ (see the arrows in the above figure). In other words, there is a relationship between the updates $\Delta(H,E')$ and $\Delta(E,E')$.

We now consider the formal derivation of this relationship. Instead of generating an expression for $\Delta(H,E')$ for some particular probabilistic interpretation, it is convenient to derive a relationship between the likelihood ratios

$$\lambda(H,E') = p(E'|H)/p(E'|\sim H) \quad (49)$$

$$\lambda(H,E) = p(E|H)/p(E|\sim H) \quad (50)$$

$$\lambda(H,\sim E) = p(\sim E|H)/p(\sim E|\sim H) \quad (51)$$

$$\lambda(E,E') = p(E'|E)/p(E'|\sim E) \quad (52)$$

The expression for any particular probabilistic interpretation can then be computed from the relation

$$\Delta(H,E) = G(\ln(\lambda(H,E))). \quad (53)$$

We begin with two expansions analogous to (48).

$$p(E'|H) = p(E'|E)p(E|H) + p(E'|\sim E)p(\sim E|H)$$

$$p(E'|\sim H) = p(E'|E)p(E|\sim H) + p(E'|\sim E)p(\sim E|\sim H)$$

Dividing the first equation by the second we get

$$\lambda(H,E') = \frac{p(E'|E)p(E|H) + p(E'|\sim E)p(\sim E|H)}{p(E'|E)p(E|\sim H) + p(E'|\sim E)p(\sim E|\sim H)} \quad (54)$$

Now we can divide both numerator and denominator by $p(E'|\sim E)$ and use $p(E|H) + p(\sim E|H) = 1$ to get

---

*Note that (48) implies

$$p(H) = p(H|E)p(E) + p(H|\sim E)p(\sim E)$$

so that the two priors, $p(H)$ and $p(E)$, cannot be specified independently. This led to problems in PROSPECTOR because experts were forced to provide priors and often did so inconsistently[14]. Instead of requiring consistent priors, PROSPECTOR accommodated the inconsistencies by using a non-Bayesian updating scheme. In this paper, we derive expressions relating probability *updates* and, in doing so, avoid these inconsistencies.

$$\lambda(H,E') = \frac{\lambda(E'|E)p(E|H) + 1 - p(E|H)}{\lambda(E'|E)p(E|\sim H) + 1 - p(E|\sim H)}. \quad (55)$$

Next, we use (50) and (51) to express $p(E|H)$ and $p(E|\sim H)$ in terms of $\lambda(H,E)$ and $\lambda(H,\sim E)$.

$$p(E|H) = \lambda(H,E) \frac{1 - \lambda(H,\sim E)}{\lambda(H,E) - \lambda(H,\sim E)}$$

$$p(E|\sim H) = \frac{1 - \lambda(H,\sim E)}{\lambda(H,E) - \lambda(H,\sim E)}$$

Combining these results with (55) we get the sequential combination function in terms of $\lambda$'s:

$$\lambda(H,E') = \frac{\lambda(E,E')\lambda(H,E)(1-\lambda(H,\sim E)) + \lambda(H,\sim E)(\lambda(H,E)-1)}{\lambda(E,E')(1-\lambda(H,\sim E)) + (\lambda(H,E)-1)} \quad (56)$$

We can derive the sequential combination function for the probabilistic interpretation $\Delta_1$ by using (53) with G given by (20). The result is

$$\Delta_1(H,E') =$$

$$\frac{-2\Delta_1(H,E)\Delta_1(H,\sim E)\Delta_1(E,E')}{(\Delta_1(H,E) - \Delta_1(H,\sim E)) - \Delta_1(E,E')(\Delta_1(H,E) + \Delta_1(H,\sim E))} \quad (57)$$

It is easy to show that (57) satisfies the axioms of sequential combination.

The sequential combination function for $\Delta_1$ is plotted in figure 7 for $\Delta_1(H,E) = .9$ and $\Delta_1(H,\sim E) = -.05, -.25,$ and $-.9$. MYCIN's sequential combination function is also plotted for comparison. The plot reveals considerable disagreement between the two functions. When $|\Delta_1(H,\sim E)|$ is near $|\Delta_1(H,E)|$, the two functions are close for $\Delta(E,E')$ greater than 0 but diverge as $\Delta(E,E')$ becomes more negative. When

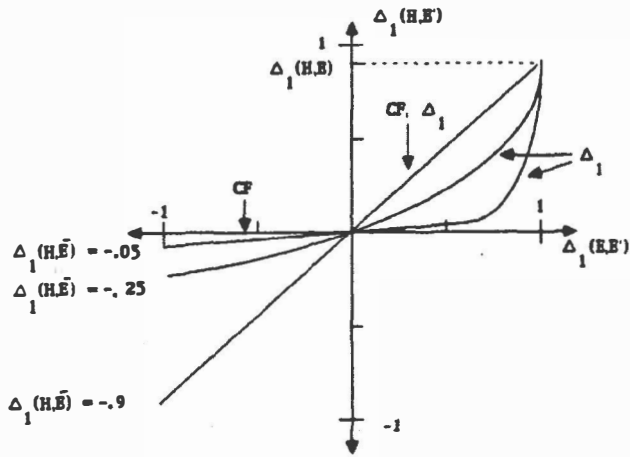

**Figure 7:** The sequential combination function for $\Delta_1$

$\Delta_1(H,\sim E)$ is near 0 (as is assumed by MYCIN), the two functions are close for $\Delta(E,E')$ less than 0 but diverge as $\Delta(E,E')$ becomes more positive.* Therefore we see that the axioms of sequential combination accommodate probabilistic interpretations whereas the particular sequential combination function used in MYCIN <u>cannot</u> have a probabilistic interpretation. Since well-understood probabilistic quantities can satisfy the intuitively appealing axioms of sequential combination, there is no reason to retain the original CF combination function.

Finally, let us consider the differences between the MYCIN and probabilistic combination functions in the example given earlier in this paper concerning the person who likes parties and solo backpacking trips. From the above discussion, it should be clear that we need to specify a certainty factor for the link

~EXTROVERT --------> DOES SOCIAL WORK

Suppose this certainty factor is -.4. In this case, if we know a person likes to go to parties and also likes solo backpacking trips, then the increase in belief that he does social work is $\Delta_1 = .2$. This is close to CF = .24 calculated earlier. In fact, the difference is due to parallel combination only since the sequential combination functions for CF and $\Delta_1$ are identical when $\Delta_1(H,E) = -\Delta_1(H,\sim E)$ and $\Delta(E,E')$ is greater than 0. However, suppose we only know that a person likes solo backpacking trips. In this case, we get CF = 0 and $\Delta_1 = -.2$ for the change in belief that he does social work. We see that the MYCIN combination function is ignoring the information contained in the link pictured above.

## Discussion

Our analysis raises two important questions.

*Given the correspondence between certainty factors and probabilities, why use certainty factors at all? Or more precisely, why retain the distinction between certainty factors and probabilities?*

One possible answer to this question is that the concept of certainty factors, or more generally, the concept of a probability update is important with respect to the user-system interaction. For example, the concept of an update can be used to assist in explanation. If a user were interested in knowing why a particular hypothesis was being considered, an expert system could display all evidence relevant to the hypothesis and the updates associated with each piece of evidence. In fact, "weight of evidence," a particular type of update mentioned earlier in this paper, is used by the GLASGOW DYSPEPSIA expert system in this way to explain its reasoning to the user[22].

Updates can also be used for entering data into an expert system. To illustrate this, consider an expert system for the diagnosis of liver disease which has in its inference net the node "patient drinks alcohol frequently" with possible values "true" and "false." Suppose the program requests that the physician enter a value for this node. The physician then presumably asks the patient whether or not he drinks frequently. Based on the patient's answer and how he answers it, the physician may not just want to enter "true" or "false" but express his uncertainty about the status of the node. It seems appropriate for the physician to relay his uncertainty as an update rather than an absolute probability. A simple argument for this can be made when the inference net has a tree structure. First note that in a tree network, prior probabilities for all non-root nodes can be calculated from the prior of the root node. In particular, the prior probabilities of the leaf or data nodes are a function of the certainty factors or updates assigned to the network by some expert. Therefore, even though a user may agree with some specified prior for the root node, his priors for data nodes may not be consistent with those of the system because he may not have the same information that is contained in the net. Thus, if the physician simply reports an absolute probability for the status of the data node after the patient reports that he does not drink frequently, the resulting updates in the net may be inaccurate due to the differences between the physician's and system's prior probability for the node. A more accurate propagation of uncertainty occurs if the physician enters an update for the data node which can then be propagated according to the prescriptions of this work. Note that the physician does not have to calculate an update "in his head." He may simply enter both his prior and posterior probabilities for the node from which the system can calculate the update.

---

*There is a simple intuitive explanation for the sharp concavity of the sequential combination function for $\Delta_1$ as $\Delta(H,\sim E)$ approaches 0. This situation corresponds to the case where $p(H)$ is near $p(H|\sim E)$. In this situation, we see from figure 6 that $p(E)$ is near 0. Since E is highly unlikely, it takes a strong positive update on E to raise its probability to a value that will alter the probability of H significantly.

### Why is MYCIN so successful?

In this paper, we discussed several restrictions that a system must satisfy before the use of certainty factors is consistent with the desiderata. However, MYCIN violates some of these restrictions. For example, the system contains sets of mutually exclusive and exhaustive hypotheses with more than two elements. In addition, certainty factors were elicited without a clear operational definition. In fact, they were elicited with the notion that they roughly corresponded to positive predictive value ($p(H|E)$), an absolute quantity[1]. However, the system performs as well as experts in the field. Perhaps detailed considerations of uncertainty is not critical to the system's performance. Indeed, a sensitivity analysis of MYCIN's knowledge base carried out by Cooper and Clancey[29] revealed that the system's performance did not depend strongly on changes in rule certainty factors. It would be interesting to study the performance of a other systems using certainty factors with respect to systematic violations of the restrictions outlined in this paper.

### Summary and conclusions

A redefinition of certainty factors in terms of the original desiderata has been proposed. The axioms of certainty factors have been examined in detail and it has been emphasized that these axioms reflect the notion that certainty factors represent a belief *update*. It has been shown that this redefinition accommodates an infinite number of probabilistic interpretations for certainty factors. These interpretations, in turn, have provided useful insights into necessary and sufficient assumptions for propagating certainty factors through an inference net. In particular, it has been shown that evidence which bears on any hypothesis in the net must be conditionally independent on the hypothesis and its negation. The analysis has made it clear that certainty factors should be elicited from experts in a precise fashion and straightforward methods for doing this have been presented. Specific improvements in the methods for sequential combination have also been presented. It is hoped that this discussion has clarified much of the mystery surrounding certainty factors.

### Acknowledgements

I would especially like to thank Eric Horvitz for many thought-provoking discussions. I would like to thank Ben Grosof for pointing out the correspondence between $\Delta_1$'s and the PROSPECTOR scoring scheme. Finally, I would like to thank Curt Langlotz, Glenn Rennels, Mark Musen, Peter Cheeseman, and Ted Shortliffe for their helpful suggestions.

### Appendix: the assumption of conditional independence

Before examining the assumption of conditional independence, we will derive some general constraints on the function F which follow directly from the basic and parallel combination axioms and and the requirements for a probabilistic interpretation. Using the definition of G, (17), and the restrictions on this function (19), the second requirement of a probabilistic interpretation (13) becomes

$$\Delta^\infty(H, E_1) < \Delta^\infty(H, E_2) \quad \text{iff} \quad p(H|E_1) < p(H|E_2).$$

Similarly, the third requirement (14) becomes

$$\Delta^\infty(H, E) = \infty \quad \text{iff} \quad (p(H|E) = 1 \text{ and } p(H) <> 0)$$

$$\Delta^\infty(H, E) = -\infty \quad \text{iff} \quad (p(H|E) = 0 \text{ and } p(H) <> 1).$$

These requirements along with the definition of F, (16), give the following restrictions on F:

F monotonic increasing

$$F(1) = \infty$$

$$F(0) = -\infty.$$

We are now ready to examine the relationship between the assumption of conditional independence and further restrictions on the function F. We begin the analysis by showing that if the function F is given by (18), then all evidence which bears directly on a hypothesis is conditionally independent of the hypothesis and its negation. As we showed earlier, when F is given by (18), the equations (16) and (31) give

$$\Delta^\infty(H, E, e) = \ln(p(E|He)/p(E|\sim He))$$

(see (33)). Since the function G has an inverse, we can write the modularity axiom (4) as follows:

$$\Delta^\infty(H, E, e) = \Delta^\infty(H, E)$$

for all e which does not entail E. Using (33) and the fact that the log function has an inverse, we get

$$\frac{p(E|He)}{p(E|\sim He)} = \frac{p(E|H)}{p(E|\sim H)} \quad (58)$$

for all e which does not entail E. Note that (58) is almost the requirement of conditional independence.

When we considered sequential combination, we saw that every rule in an inference net of the form IF E THEN H must be associated with a certainty factor $\Delta(H, \sim E)$, the change in belief of H given the *absence* of E, as well as a certainty factor $\Delta(H, E)$. Given this, we get the companion requirement of (58)

$$\frac{p(\sim E|He)}{p(\sim E|\sim He)} = \frac{p(\sim E|H)}{p(\sim E|\sim H)} \quad (59)$$

for all e which does not entail $\sim E$. From (58) and (59) it follows that either

$$p(E|He) = p(E|\sim He) \quad (60)$$

for all e which does not entail E or its negation, or that

$$p(E|He) = p(E|H) \text{ and } p(E|\sim He) = p(E|\sim H) \quad (61)$$

for all e which does not entail E or its negation. (60) implies $\Delta(H, E) = 0$ which means that E has no affect on H. Therefore, when F is given by (18), evidence which has any bearing on a hypothesis must be conditionally independent on the hypothesis and its negation.

The converse is also true. That is, if we assume conditional independence, F is essentially given by (18). To see this, first note that since F is monotonic increasing, we can write

$\Delta^\infty(H,E,e)$

$= \ln(f'(p(H|Ee)/1-p(H|Ee))) - \ln(f'(p(H|e)/1-p(H|e)))$ (62)

$= \ln(f'(O(H|Ee))/f'(O(H|e)))$

where f' is a monotonically increasing, smooth function. In conjunction with the axiom of modularity, (4), this gives

$f'(O(H|Ee))/f'(O(H|e)) = f'(O(H|E))/f'(O(H))$ (63)

for all e which does not entail E. Now suppose all evidence for some hypothesis is conditionally independent on the hypothesis and its negation. Using this assumption and (31) we get

$O(H|Ee)/O(H|e) = O(H|E)/O(H)$ (64)

for all e which does not entail E. (63) and (64) together imply that

$f'(x) = A x^\alpha \quad A, \alpha > 0$

which means

$F(x) = \alpha \ln(x/1-x) \quad \alpha > 0$ (65)

Since the factor $\alpha$ can be incorporated into the definition of G, we see that the assumption of conditional independence amounts to choosing F to be the function given by (18).

From the above analysis we see that the dependencies among evidence given hypothesis is directly related to the choice for F. In particular, if we assume that F is not given by (65), then we must assume that *all* evidence which bears directly on a hypothesis is conditionally dependent in some *uniform* way determined by the exact choice for F. The assumption of uniform conditional dependence seems far less attractive than the assumption of conditional independence. Therefore, we add the conditional independence assumption to the requirements for a probabilistic interpretation keeping in mind that there is no other reasonable alternative.

As mentioned in the text, it was the intention of the creators of certainty factors to require "modularity" since this would facilitate the construction of the knowledge base. The above analysis makes it clear that "modularity" corresponds to conditional independence in the probabilistic sense.